\documentclass[letterpaper]{article} 
\usepackage{aaai23}  
\usepackage{times}  
\usepackage{helvet}  
\usepackage{courier}  
\usepackage[hyphens]{url}  
\usepackage{graphicx} 
\usepackage{natbib}  
\usepackage{caption} 
\usepackage{algorithm}
\usepackage{algorithmic}
\usepackage{amsmath}

\usepackage{newfloat}
\usepackage{listings}
\DeclareCaptionStyle{ruled}{labelfont=normalfont,labelsep=colon,strut=off} 
\floatstyle{ruled}
\newfloat{listing}{tb}{lst}{}
\floatname{listing}{Listing}
\newcommand{\ie}{\textit{i}.\textit{e}., }
\newcommand{\eg}{\textit{e}.\textit{g}. }

\title{Parametric Surface Constrained Upsampler Network for Point Cloud}
\author {
    Pingping Cai,
    Zhenyao Wu,
    Xinyi Wu,
    Song Wang
}
\affiliations {
    University of South Carolina, USA \\

    \{pcai,zhenyao,xinyiw\}@email.sc.edu, songwang@cec.sc.edu
}

\usepackage{bibentry}

\begin{document}
\maketitle

\begin{abstract}
Designing a point cloud upsampler, which aims to generate a clean and dense point cloud given a sparse point representation, is a fundamental and challenging problem in computer vision.
A line of attempts achieves this goal by establishing a point-to-point mapping function via deep neural networks.
However, these approaches are prone to produce outlier points due to the lack of explicit surface-level constraints.
To solve this problem, we introduce a novel surface regularizer into the upsampler network by forcing the neural network to learn the underlying parametric surface represented by bicubic functions and rotation functions, 
where the new generated points are then constrained on the underlying surface.
These designs are integrated into two different networks for two tasks that take advantages of upsampling layers -- point cloud upsampling and point cloud completion for evaluation.
The state-of-the-art experimental results on both tasks demonstrate the effectiveness of the proposed method.
The code is available at https://github.com/corecai163/PSCU.
\end{abstract}

\section{Introduction}
\label{intro}
\begin{figure}
	\centering
	\includegraphics[width=0.470\textwidth]{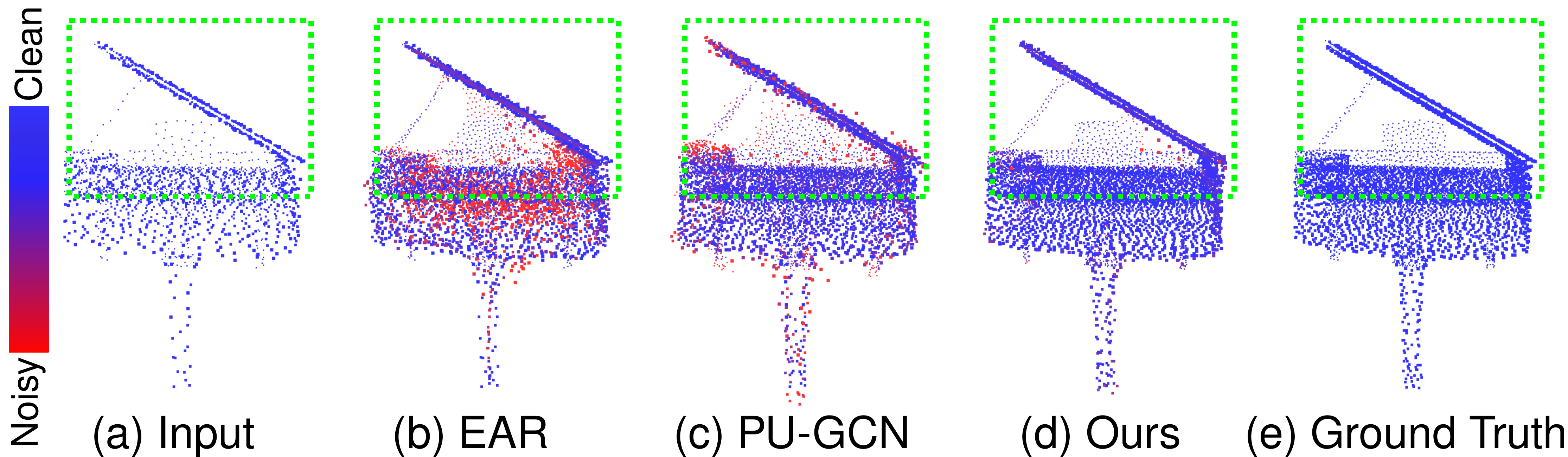}
	\caption{The double-layer lid can be clearly upsampled by our method,
   while the traditional EAR \cite{ear} fails to distinguish the two nearby layers and the previous SOTA method PU-GCN \cite{PUGCN} generates many noisy and outlier points. Please zoom in for more details.
   }
	\label{fig:mlp_fail}
\end{figure}

Point cloud is an efficient data structure to represent 3D objects. But, due to the limitation of sensors, the collected point clouds are usually sparse and incomplete. Therefore, point cloud upsampling \cite{pugan,PUGCN,punet,pugeo,Wang2019PatchBasedP3} is introduced to generate denser point clouds for better scene representation, which benefits many computer vision applications such as autonomous driving \cite{pcdForAutoDrive,pcdForAutoDrive2}, 3D object classification \cite{Li_2020_CVPR,pointnetplusplus}, semantic segmentation \cite{pcd_seg,Landrieu_2018_CVPR,Engelmann_2017_ICCV} and robotics \cite{pcdForRobot}. 
For point cloud upsampling, it is expected that the generated dense points can well represent the shape and surface underlying the point cloud.
However, obtaining such property is challenging, and even previous state-of-the-art (SOTA) methods may generate many noisy and outlier points (see Figure \ref{fig:mlp_fail}).

Many sophisticated methods have been proposed to solve this challenging problem.
Traditional optimization based methods \cite{traditional_1,traditional_2,ear,traditional_4,traditional_5} rely on geometric priors to upsample the point cloud but often fail when the input is complex and have been outperformed by recent deep learning methods.
These well-designed deep learning methods can be 
further divided into two categories based on their adopted upsampler:
1) feature-based upsampling methods and 2) folding-based upsampling methods.
The feature-based methods \cite{SFA,PF-Net,wu2019point,MetaPUAA,PUGCN,punet,snowflakenet} first extract shape features from input points and expand the number of points by upscaling the shape features in the feature space. These upsampled features are then fed into a coordinate regression block (MLPs) to predict their coordinates. 
A point-wise loss function, \textit{i.e.,} Chamfer Distance \cite{FanSG17}, is usually used to train the network.
However, this loss function only measures the point-wise distance and can not measure the difference of underlying surfaces between point clouds.
As a result, these methods often fail to generate points that are located accurately on the underlying surfaces.
Folding-based methods \cite{foldingnet,pcn,msn,spu,pc2pu,luo2020differentiable} expand the number of points by introducing a predefined 2D grid for each point and then concatenating them with shape features to regress for their coordinates via MLPs -- they
can be viewed as a mimic of a morphing/surface function that transforms the 2D grid to target surfaces.
While they attempt to preserve a better surface structure,
they can only learn an overfitted point-point mapping with point-wise loss functions \cite{Deep_geometric_prior}, 
but not the accurate representation of the surface.
To achieve a better surface representation,
PUGeo-Net \cite{pugeo} introduced a parameterized method that incorporates discrete differential geometry into network design, where it models the small surface around each input point via the first and second fundamental forms \cite{differential_geometry} to generate the upsampled points. However, this method relies heavily on the correctness of point normals and needs additional ground truth point normals to train the network. Such normals are not directly available in many point cloud datasets. 
Besides, it still uses MLPs to predict the point displacement and lacks explicit surface-level constraints.
Also, it follows the patch-based upsampling pipeline, which does not consider the smoothness between two adjacent patches.

We extend PUGeo-Net by using explicit parametric functions to model the local surface for each input point without the required supervision of point normal information.
Besides, instead of dividing the input point cloud into multiple patches, we directly upsample the entire input to avoid the discrepancy between adjacent patches.
Specifically, we design a novel parametric surface constrained upsampler network that can extract the spatial surface features from discrete and unordered input points, predict explicit bicubic functions and rotation functions to express the underlying surfaces, and constrain the new generated points on these surfaces.
To further improve its performance, we also introduce a displacement loss for generating better parametric functions.
The proposed upsampler can be also used for other related tasks such as point cloud completion.

%
We evaluate our proposed method in both point cloud upsampling and completion tasks on three standard datasets, PU1K, KITTI, and ShapeNet-PCN.
The experiment results demonstrate that by using the proposed surface constrained upsampler,
we can achieve new SOTA results by outperforming previous methods.
The main contributions of this paper are as follows. 
\begin{enumerate}
\item 
We propose a novel surface-level constraint that uses parametric surfaces as regularizers to ensure the smoothness of the upsampled point clouds. 
\item We design a new parametric surface constrained upsampler, that estimates the surface parametric functions from input points and generates new points on each surface.
\item We evaluate the proposed upsampler on both point cloud upsampling and point cloud completion tasks, and our proposed method achieves new SOTA performance. 
\end{enumerate}

\section{Related Work}
\label{related}
\subsection{Point Cloud Upsampling}

\citeauthor{punet} \shortcite{punet} proposed the first deep learning based point cloud upsampling algorithm PU-Net at patch level. For each patch, multilevel features are extracted at each input point and expanded  via a multibranch convolution unit in the feature space. Then, these expanded features are reconstructed for upsampled point cloud via an MLP based coordinate regression block.
However, when the upsampling rate is high, \eg $\times 16$, it needs to define multiple branches of convolution layers, which is quite inefficient.
To address this issue, 
\citeauthor{Wang2019PatchBasedP3} \shortcite{Wang2019PatchBasedP3} proposed 3PU, a multistep patch-based method to progressively upsample the points by breaking the upsampling task into multiple $\times 2$ small tasks, each solved by a sub-network that focuses only on a particular level of detail.
Each sub-network has the same structure with the feature extractor unit, feature expansion unit, and an MLP to regress for the point coordinates.
To consider more spatial information among neighboring points,
\citeauthor{PUGCN} \shortcite{PUGCN} proposed PU-GCN by introducing
a multi-scale Inception DenseGCN feature extractor
to extract the spatial features among nearby points
and another graph convolution network to better encode local point information from its neighbors.
Although PU-GCN achieves the SOTA result by utilizing spatial information, it relies on an MLP-based coordinate regression module which often fails to learn the accurate representation of the surface.

\subsection{Point Cloud Completion}
Point cloud completion can be considered as a more challenging version of upsampling where the input point cloud is sparse and incomplete. It aims to not only upsample the input point cloud but also infer the missing underlying shapes and surfaces \cite{vrpcn,pcn,grnet,snowflakenet,pmp,DeepPCD}. 

\citeauthor{pcn} \shortcite{pcn} proposed PCN, a novel coarse to fine point cloud completion network under an encoder-decoder framework. It first extracts the global shape code that describes the coarse shape of the input points using a PointNet \cite{pointnet} feature extractor. Then, a coarse but complete point cloud is generated from the global shape code and fed into a folding-based upsampling block to generate the dense point cloud.
However, PCN's feature extractor can not extract sufficient structural information from input point clouds and may not well capture the geometric detail.
To this end, 
\citeauthor{grnet} \shortcite{grnet} proposed GRNet, using 3D grids as intermediate representations to regularize unordered point clouds and 3D convolution networks to explore the structural context and infer the underlying shape of the input point clouds.
However, the resolution of intermediate point clouds is limited by the size of 3D feature maps, making it hard to reveal fine local geometric details. As a result, it still needs additional MLPs to refine and upsample the point clouds.
To preserve more local geometric details, 
\citeauthor{snowflakenet} \shortcite{snowflakenet} designed SnowFlakeNet. It first extracts the global shape code from input point clouds using PointNet++ \cite{pointnetplusplus} and Point Transformer \cite{point_transformer} blocks, which are fed into a seed generator to generate a course but complete point cloud. The coarse point cloud is then upsampled to a denser one via SnowFlake Point Deconvolution blocks. These blocks consist of 1D deconvolution layers to upsample the points in the feature space and skip-transformers to preserve better structure and shape.
Although SnowFlakeNet achieves the state-of-the-art result on point cloud completion tasks, it still relies on an MLP to regress for the coordinates of each upsampled point without constraining the upsampled points to be on the underlying surfaces accurately.

\section{Parametric Surface Constrained Upsampler}
Designing an upsampler for point clouds is nontrivial due to the difficulty of inferring accurate underlying surfaces from discrete points and elegantly constraining the upsampled points on these surfaces.
We solve these challenges by forcing the network to predict an explicit representation of the underlying surface via parametric surface functions and then generate points directly on the surface via the predicted parametric functions.

\begin{figure}[bt]
	\centering
	\includegraphics[width=0.45\textwidth]{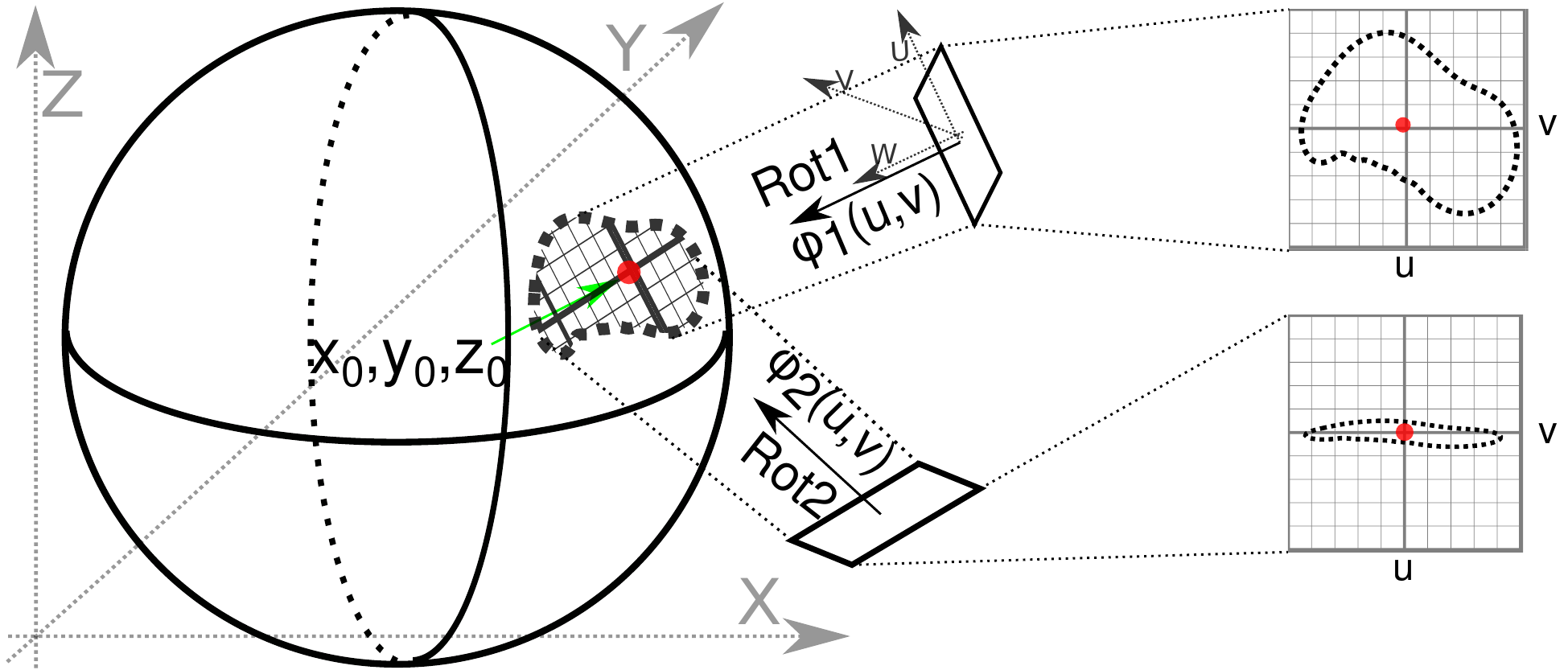}
	\caption{An example to describe the surface of a small region by the parametric functions. There exist multiple projections and surface functions to express this small region.}
	\label{fig:background}
\end{figure}

\subsection{Parametric Surfaces} 
\label{manifold_background}
Ideally, given points on the 2D plane X-Y, we can map them into 3D space via an explicit function $\Phi(x,y) \rightarrow (x,y,z)$. However, if the target surface is perpendicular to the X-Y plane, \ie $X = 0$, we cannot calculate the coordinates $z$ for $x \neq 0$, limiting its representation ability. To solve this problem, we use local coordinate systems with rotations to improve the representation ability with better projection planes.
Figure \ref{fig:background} shows an example of local parametric surfaces around the coordinate $(x_0,y_0,z_0)$. 
In particular, we define the parametric surface function as 
$ Rot(\phi(u,v))+(x_0,y_0,z_0)$
, where $\phi(u,v) \rightarrow (u,v,w)$ is the surface function that maps
points in 2D projection planes to 3D surfaces on a local coordinate system U-V-W, and
$Rot(u,v,w) \rightarrow (x,y,z)$ is the rotation function with the rotation center $(u=0,v=0,w=0)$
to rotate local coordinates into the global X-Y-Z coordinate system.
Mathematically, given these two functions, we can generate an arbitrary number of points on the local surface.
Thus, we introduce this idea into the design of the surface constrained upsampler network, \textit{i.e.}, we propose to learn and utilize the surface function $\phi$ and the rotation function $Rot$ to constrain the upsampled points.

\textbf{Explicit Surface Function}
Previous methods used the concatenation-based folding technique to model the surface function via MLPs \cite{foldingnet,pcn,pugeo,luo2020differentiable}.
However, as mentioned in the introduction,
such methods only learn an overfitted point-point mapping and induce a bottleneck that limits its capacity to represent different 3D surfaces.
Different from these methods,
we propose to use the explicit polynomial function to model the underlying surface, which can be expressed as:
\begin{equation}
w = \phi(u,v) = \sum_{i}\sum_{j}a_{ij} u^i v^{j},
\end{equation}
where $a_{ij}$ is the coefficient predicted via neural networks.
With different combinations of coefficients, we can easily express different shapes.
For simplicity, we use the bicubic function, which is widely used and strong enough to express common shapes, to design our network:
\begin{equation}
   w = \phi(u,v) = a_1 + a_2 u + a_3 u^2 + ... + a_{16} u^3v^3,
\label{eq:pj} 
\end{equation}

\textbf{Rotation Function}
After generating upsampled points on the local surface, 
we need to rotate them from their local coordinate systems into the global coordinate system.
To achieve this, we model the rotation function as follows:
\begin{equation}
   \begin{pmatrix} x \\ y \\z \end{pmatrix}
= Rot(u,v,w) = \begin{pmatrix} r_1 & r_2 & r_3 \\
r_4 & r_5 & r_6\\ r_7 & r_8 & r_9 \end{pmatrix}
\begin{pmatrix} u \\ v \\w \end{pmatrix},
\label{eq:rot} 
\end{equation}
where $[r_1,...,r_9]$ are the elements of a rotation matrix $R$ predicted by neural networks.
To ensure that the rotation matrix follows the correct principle of $R^T*R == R*R^T == I$, we use the 6d representation proposed in \cite{Rotation_Representations}, which shows a good continuous property and can be decoded into a $3 \times 3$ matrix.

\subsection{Network Design}
\label{network_design}

\begin{figure}[bt]
	\centering
	\includegraphics[width=0.47\textwidth]{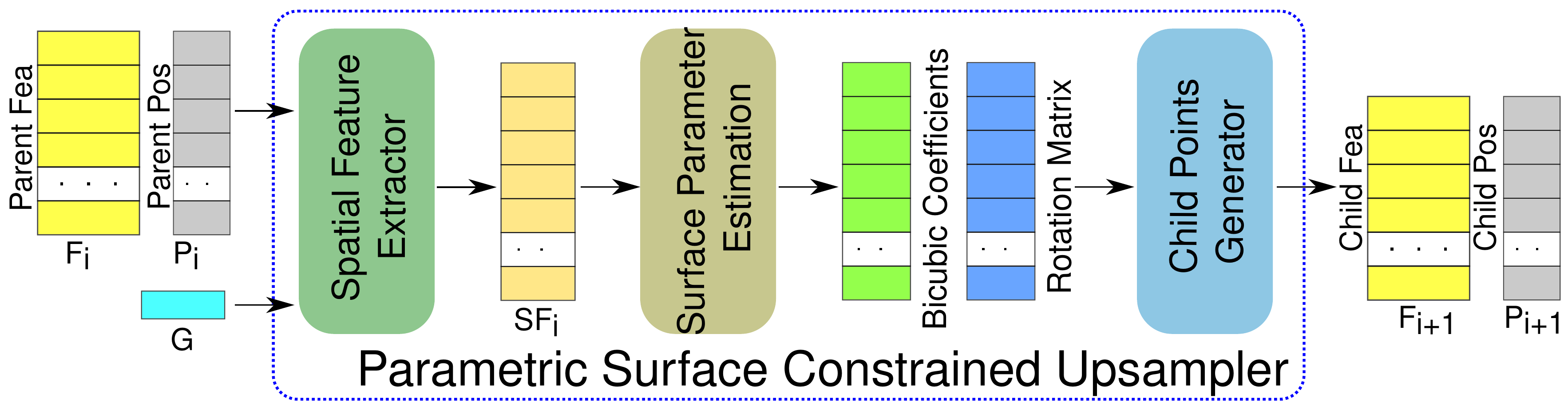}
	\caption{The general structure of the parametric surface constrained upsampler. Note that due to the page limitation, the detailed network structure is in the Supplementary.}
	\label{fig:MDConv}
\end{figure}

However, designing a special network that can predict accurate surface parameters and seamlessly integrate them into the upsampling procedure is challenging.
For convenience, we refer to the input points as parent points and the upsampled points as child points.
Our idea is that each parent point will split and generate multiple child points that lie on the local surface and cover the entire surface as much as possible.
To achieve this, we design the parametric surface constrained upsampler network that contains 3 major parts: (\romannumeral1) Spatial Feature Extractor, (\romannumeral2) Surface Parameter Estimation, and (\romannumeral3) Child Points Generator, aiming at extracting the local geometric shape from unordered parent points, predicting the explicit surface parameters around each parent point, and generating child points on parametric surfaces, respectively. 
Each of them will be described in detail in the following subsections.
Figure \ref{fig:MDConv} shows the general structure of the proposed upsampling network, and
note that the detailed network structure is provided in the Supplementary.
The proposed upsampler module requires three necessary inputs: parent coordinates $P_{i} \in \mathbf{R}^{N \times 3}$, parent features $F_{i} \in \mathbf{R}^{N \times C_1}$, and the global shape code $G \in \mathbf{R}^{1 \times C_2}$ that encodes the global shape of the input point cloud.
It aims to upsample them $m$ times and generates child points with coordinates $P_{i+1} \in \mathbf{R}^{mN \times 3}$ and features $F_{i+1} \in \mathbf{R}^{mN \times C_1}$.


\subsubsection{\romannumeral1. Spatial Feature Extractor}

Making the network aware of the local geometric information around each parent point
is a key step in predicting accurate shapes. To achieve this, we design a Spatial Feature Extractor (SFE) that can aggregate the positions and features of the parent's $K$ nearby points to extract the local spatial information. 
Especially, the SFE extracts the local spatial features $SF_{i} \in \mathbf{R}^{N \times C_1}$ from input parent features $F_{i}$, parent positions $P_{i}$, and global shape code $G$,
which is defined as 
\begin{equation}
SF_{i} = \mathtt{Aggregate}_K(P_{i},\mathtt{NN}_s(F_{i},G)),
\end{equation}
where $\mathtt{Aggregate}_K$ is the point transformer introduced in \cite{point_transformer} to aggregate the context of $K$ nearest points, 
and $\mathtt{NN}_s$ is the neural network used to combine the parent features $F_{i}$ with global shape code $G$.

\subsubsection{\romannumeral2. Surface Parameter Estimation}

%

Then, to accurately represent the underlying surface around each parent point, we propose to estimate the explicit parameters of the surface function $\phi$ and the rotation function $Rot$ as mentioned before.
One simple way is to directly predict their parameters from local spatial features $SF_{i}$.
However, $SF_{i}$ only contains the local shape information, and the global smoothness of these local shapes is not guaranteed. 
Thus, we incorporate the global shape code $G$ to smooth them.
Especially, these parameters can be predicted via:
\begin{equation}
    a = \mathtt{NN}_a(G,SF_{i})),
\end{equation}
\begin{equation}
    r = \mathtt{NN}_r(G,SF_{i})),
\end{equation}
where $\mathtt{NN}_a$ and $\mathtt{NN}_r$ are neural networks and $a$ and $r$ are predicted coefficients of the bicubic function and the rotation matrix.


\begin{figure}[bt]
	\centering
	\includegraphics[width=0.47\textwidth]{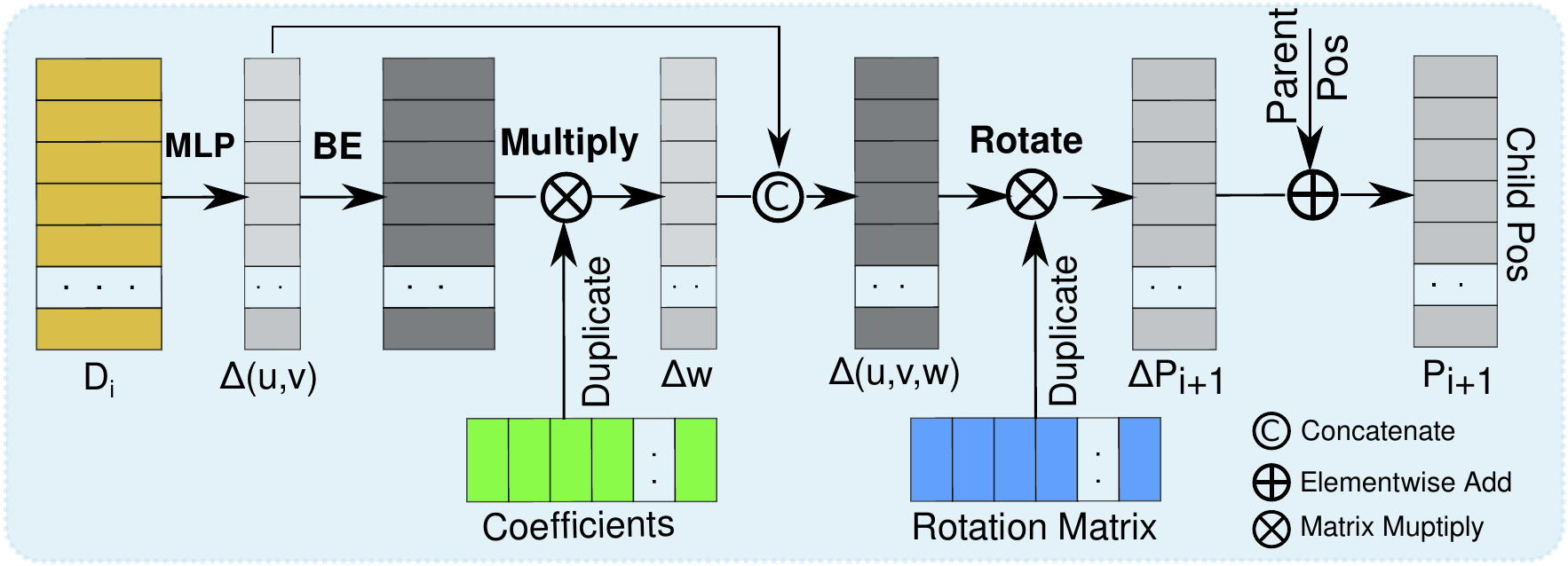}
	\caption{The network architecture for generating child points on surface via predicted surface parameters.}
	\label{fig:child_points}
\end{figure}

\subsubsection{\romannumeral3. Generating Child Points on Surface}
Finally, our objective is to generate the child features $F_{i+1}$ and the child positions $P_{i+1}$ on the predicted parametric surface.
Unlike previous methods \cite{snowflakenet,PUGCN,punet}, where
they reconstruct the 3D child coordinates directly from the child features through MLPs, 
we design a network that smoothly integrates the predicted surface parameters into child point generation.
Especially, we first predict the displacement of the child $(\Delta u, \Delta v)$ in the projection plane and then lift them into 3D spaces using predicted parametric functions.

To implement this, we first generate the relative displacement of the child features $D_{i} \in  R^{mN \times C_1}$ \textit{w.r.t} to their parents'
through the 1D deconvolution layer, which can easily generate different numbers of child features by setting different kernel sizes and strides.
The displacement feature $D_{i}$ is used to predict the coordinate displacement of the child $(\Delta u, \Delta v)$ using an MLP. 
Specifically, $(\Delta u, \Delta v) = \mathtt{MLP}(D_{i})$.
Next, based on Equation \eqref{eq:pj}, we can calculate their embedded values [$1$,$\Delta u$,$\Delta u^2 $,...,$\Delta u^3 \Delta v^3$] via a Bicubic Embedding ($\mathtt{BE}$) block and multiply them with predicted bicubic coefficients to generate the coordinate displacement $\Delta{w}$.
We then transit them into X-Y-Z coordinate system 
via the predicted rotation matrix and get the child displacements
$\Delta P_{i+1} = (\Delta x,\Delta y,\Delta z)$,
which will be added with
their parent positions $P_i$ to obtain the final position of the child points $P_{i+1}$.
Figure \ref{fig:child_points} shows the corresponding network architecture to generate the child position in the parametric surface. 
After obtaining the position of the child points, we feed $D_i$ into another MLP layer and add the output with their parent feature to get the child features $F_{i+1}$.

\subsection{Loss Function}
\label{loss}
To train our network, we first use Chamfer Distance as a loss function for each upsampling block, which measures the point-wise distance between the predicted point cloud and ground truth.
However, we notice that given a small area of parametric surface there exist multiple projection planes with different surface functions and rotation functions.
Not all of them can describe this small surface correctly and efficiently.
For example, in Figure \ref{fig:background}, if the projection plane is perpendicular to the surface (parallel to the surface normal), it is difficult to find a good surface parametric function.

Ideally, we aim to select a projection plane that is perpendicular to the normal of each surface/parent point.
As our network does not have ground truth normal information,
we borrow the idea from the unsupervised principal component analysis algorithm \cite{PCA} to select a better projection plane, 
where it aims to find a projection plane that 
can maximize the covariance matrix of $(\Delta u, \Delta v)$, \textit{a.k.a}, minimize the covariance matrix of $\Delta w$ given 3D points in a small region.
Thus, inspired by this we add a constraint to our network by introducing the displacement loss as follows:
\begin{equation}
    \mathcal L_d = ||\Delta w||_2^2. 
\end{equation}
In summary, our final loss function is defined as
$ \mathcal L = \mathcal L_{\textit{CD}} + \lambda \mathcal L_d $,
where $\lambda$ is a hyperparameter that balances the weight of the chamfer loss and the displacement loss.

\section{Experiments}
To validate the effectiveness of the proposed upsampler,
we first evaluate it on the PU1K dataset and conduct ablation studies to verify the effectiveness of our network. We also apply our method to the real collected LiDAR point cloud KITTI dataset \cite{data_kitti}.
Then, we further present the capability of the proposed method on a more challenging point cloud completion task on the ShapeNet-PCN dataset.

\subsection{Point Cloud Upsampling} \label{exp:1}

\noindent
\textbf{PU1K:}
The PU1K dataset is first introduced in PU-GCN \cite{PUGCN} for point cloud upsampling. It consists of 1,147 3D models, which are collected from PU-GAN \cite{pugan} and ShapeNet dataset \cite{chang2015shapenet} to cover various shape complexities and diversity.
The input point cloud is sparse but complete with 2,048 points, and the ground truth point cloud is 4 times denser with 8,192 points.
We follow the same train/test splitting strategy in PU-GCN
with 1,020 training samples and 127 testing samples.
Note that unlike previous patch-based methods \cite{PUGCN,MetaPUAA,punet,PUTrans},
our training data are entire point clouds generated from ground truth meshes by poisson disk sampling \cite{poisson_disk}. For testing, we directly use the test data given by PU-GCN for a fair comparison.

\begin{figure}[bt]
	\centering
	\includegraphics[width=0.47\textwidth]{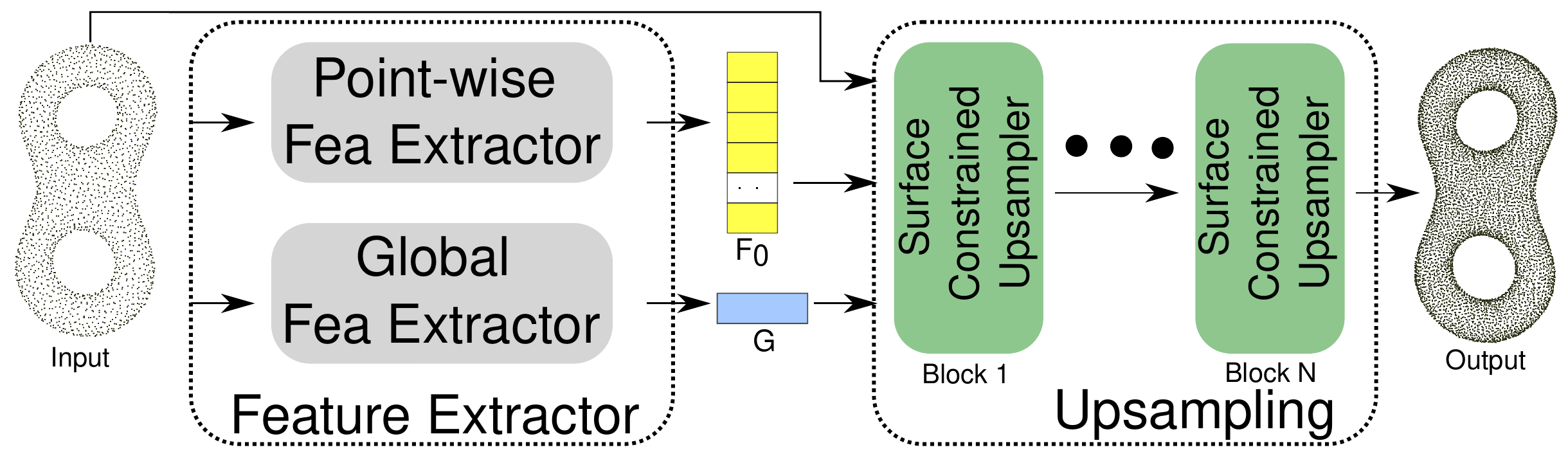}
	\caption{Point cloud upsampling with surface constrained upsampler. We can stack multiple upsampling blocks to achieve a higher upsampling ratio.}
	\label{fig:overview}
\end{figure}
\noindent
\textbf{Network Structure:}
Figure \ref{fig:overview} shows the network architecture for the point cloud upsampling task.
To generate the required inputs for the proposed upsampler, we use a feature extractor to capture both point-wise features and global features from sparse points.
Especially, it consists of two parts: a point-wise feature extractor, which is an MLP that 
maps input points into the feature space, and a global feature extractor, which consists of a PointNet++ backbone \cite{pointnetplusplus} with point transformers \cite{point_transformer} to
incorporate both the local and global shape contexts.
The outputs of the feature extractor block are the point-wise features $F_{0}$ and the global shape code $G$.
Note that the design of feature extractor is not the major contribution of this paper and we can exploit any other suitable networks.
We then feed these outputs along with the original position of the points $P_{0}$ into stacks of upsampling blocks to generate denser point clouds.
To upsample the point cloud 4 times,
we arbitrarily set two upsampler blocks with upscale ratios of 1 and 4, respectively. 
Note that other combinations of upscale ratios and the number of upsampler blocks are also feasible. 

\noindent
\textbf{Evaluation Metrics:}
We use three widely adopted metrics in previous work to evaluate our performance:
Chamfer Distance (CD), Point-to-Surface Distance (P2F) \textit{w.r.t} ground truth meshes, and Uniformity Score \cite{pugan}.
For these metrics, a smaller value means better performance.

\noindent
\textbf{Training Detail:} 
To train this network, we use 2 Tesla V100 GPUs. 
We set the batch size to 16 and the total epoch number to 150. 
Besides, we use Adam as the optimization function with a learning rate of
0.0005 at the beginning, and we decrease the learning rate by a factor of 0.5 every 50 epochs.

\subsubsection{Experiment Results:}

Table \ref{tab:pu1k} shows the quantitative upsampling results on the PU1K dataset.
We find that our algorithm achieves
the best performance over all its counterparts with large improvements.
In particular, compared to the previous SOTA algorithm, PU-GCN,
the proposed algorithm reduces the average CD from $1.151 \times 10^{-4}$ to $0.886 \times 10^{-4}$.
Besides, the average P2F also reduces more than half from $2.504 \times 10^{-3}$
to $1.091 \times 10^{-3}$, which statically proves that our generated child points preserve better surface shapes and locate closer to the ground truth surfaces.
What's more, the proposed method also obtains better uniformity scores than previous SOTA algorithms.
\begin{figure*}[bt]
	\centering
	\includegraphics[width=0.89\textwidth]{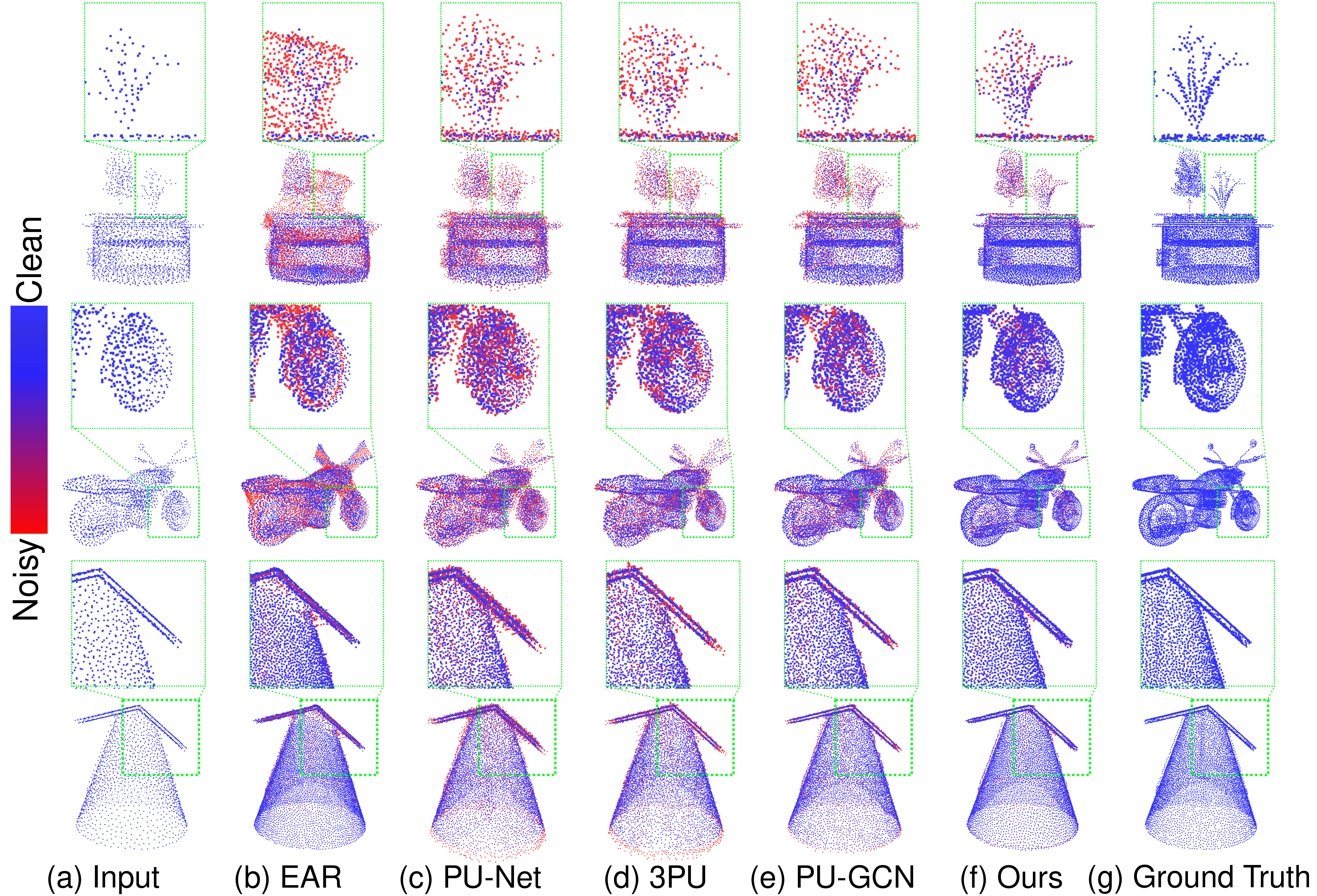}
	\caption{Visualization of upsampling results with different algorithms (EAR, PU-Net, 3PU, PUGCN, and Ours). We see that our method produces the best results, generating smooth borders and preserving fine-grained local details.}
	\label{fig:pu1k}
\end{figure*}
Next, to intuitively show the performance, we visually check the upsampling outputs of our method
and compare them with the outputs of other algorithms.
Figure \ref{fig:pu1k} shows the visual results of different
algorithms.
We see that our proposed method can produce point clouds with much better shape quality and fewer off-the-surface points.
Apparently, both the quantitative and visual results prove the superiority of the proposed
network.
\begin{table}[bt]
	\centering
	\begin{tabular}{p{0.074\textwidth}|p{0.05\textwidth}p{0.05\textwidth}p{0.03\textwidth}p{0.03\textwidth}p{0.03\textwidth}p{0.03\textwidth}p{0.03\textwidth}}
			\hline\noalign{}
    Method & CD  & P2F & \multicolumn{4}{c}{Uniformity $\times (10^{-3}$)}\\
     & ($\times 10^{-4}$) & ($\times 10^{-3}$) &0.4\% & 0.6\% & 0.8\% & 1.0\% \\
     \hline
    EAR & 1.449 & 3.314 &  1.82 & 3.68 & 6.51 & 9.92 \\
    PU-Net & 1.751 & 4.847  & 2.07 & 4.24 & 7.54 & 11.78\\
    3PU & 1.461 & 3.559 &  1.99 & 4.12 & 7.23 & 11.04\\
    PU-GCN & 1.151 & 2.504  & 1.95  &3.97 & 6.83 & 10.63\\
    Ours &\bf{0.886} & \bf{1.091}  &\bf{1.40} & \bf{2.85} & \bf{5.06} & \bf{7.95}\\
    \hline
    \end{tabular}
  \caption{Quantitative upsampling results compared to previous SOTA algorithms. The uniformity score is estimated in the local area of different percentages of radii.}
  \label{tab:pu1k}
\end{table}

\begin{table}[bt]
	\centering
		\begin{tabular}{c|cc}
		\hline
 			Components & CD ($\times 10^{-4}$)   \\
 			\hline
			w/o Spatial Feature & 1.293 & \\
			w/o Parametric Function & 0.967 \\
			w/o Displacement Loss & 0.893  \\
			\hline
			Full & \bf{0.886} \\
			\hline
		\end{tabular}
  \caption{Ablation studies of the Parametric Surface Constrained Upsampler on the PU1K dataset.}
  \label{tab:ablation}
\end{table}

\subsection{Ablation Study} \label{exp:2}
We then perform ablation studies to figure out which part of the proposed upsampling network contributes the most to its performance. Table \ref{tab:ablation} summarizes all experiment results.

\textbf{Spatial Feature:}
Intuitively, the local spatial information is crucial to predict accurate local surfaces.
Thus, we first examine its importance by removing the point transformer in the Spatial Feature Extractor, which is designed to aggregate features of $K$ nearest neighbors for each point. 
In Table \ref{tab:ablation}, we see that after removing the point transformer the performance
drops to $1.239 \times 10^{-4}$ with a huge gap, which justifies our intuition.

\textbf{Parametric Function:}
As we explicitly model the parametric surface via the bicubic function and the rotation function, one might ask about the effectiveness of this explicit representation compared to folding-based MLPs \cite{pcn,msn,foldingnet,pointSA}.
To this end, we substitute the explicit surface function with an MLP-based folding layer that takes the displacement of the child points $(\Delta u, \Delta v)$ as inputs and outputs its global displacement $(\Delta x,\Delta y,\Delta z)$.
In Table \ref{tab:ablation}, we see that after using MLPs to model surface functions, performance decreases and CD increases from $0.886 \times 10^{-4}$ to $0.967 \times 10^{-4}$ with a gap of $0.081 \times 10^{-4}$. 
This gap illustrates the superiority of parametric surface functions in representing better underlying surfaces compared to MLPs.

\textbf{Displacement Loss:}
Since we introduce an additional displacement loss to select a better projection plane,
we then illustrate its effect on training the proposed upsampler by removing this loss.
We see that without displacement loss, the performance decreases slightly from $0.886 \times 10^{-4}$ to $0.893 \times 10^{-4}$. 
This slight performance drop fits our intuition because there exist multiple choices for the projection planes, and it contributes the least to the performance of the proposed upsampler.


\subsubsection{Robustness to Noise}
\begin{figure}[bt]
	\centering
	\includegraphics[width=0.47\textwidth]{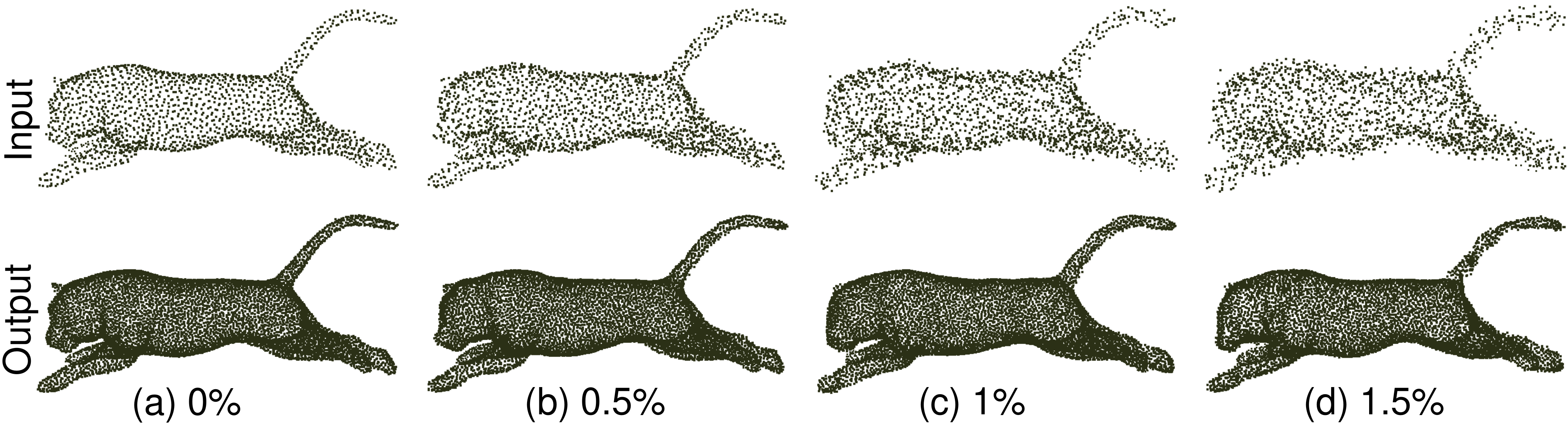}
	\caption{Visualization of upsampling results with different levels of noise. Our method still can preserve good underlying shapes with noisy inputs.}
	\label{fig:noise}
\end{figure}

Another concern might be the performance of the proposed method with noisy inputs.
Therefore, we add a small perturbation to each input point with a Gaussian distribution to synthesize noise, retrain and test the robustness of our method at different noise levels.
Table \ref{tab:noise} and Figure \ref{fig:noise} show the quantitative and visual results at different levels of perturbations.
\begin{table}[bt]
	\centering
		\begin{tabular}{ccccc}
			\hline
			Noise & $0\%$ & $0.5\%$ & $1\%$ & $1.5\%$ \\
			\hline
			CD ($\times 10^{-4}$)  & 0.886 & 1.053  & 1.280 & 1.599 \\
			\hline
		\end{tabular}
  \caption{Performance under different noise perturbations.}
  \label{tab:noise}
\end{table}
Intuitively with noise points, it becomes more difficult to infer the accurate surfaces.
We see that under a small perturbation such as 0.5\%, our method still achieves a promising result. 
Even with a 1.5\% perturbation, the upsampled points still tend to preserve a smooth shape with little distortion. 

\begin{figure}[bt]
	\centering
	\includegraphics[width=0.47\textwidth]{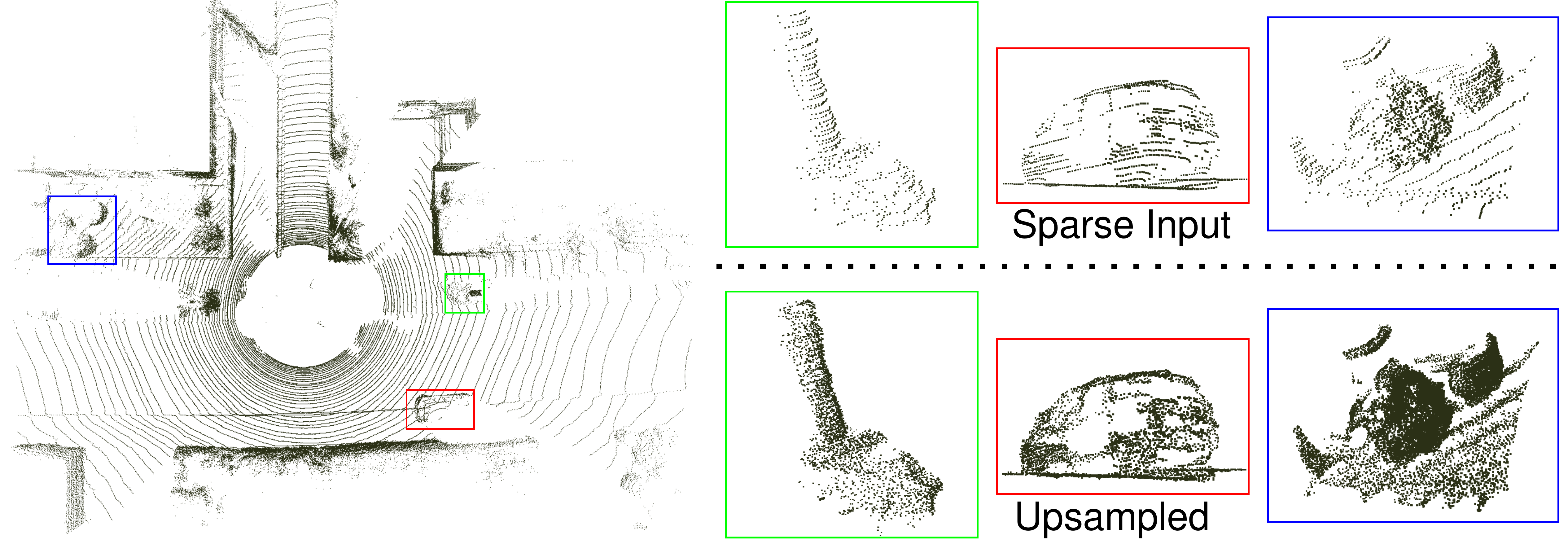}
	\caption{4 times upsampling result on a real-world point cloud from KITTI dataset.}
	\label{fig:kitti}
\end{figure}
\subsubsection{Result on Real World Data}
Finally, we further show the performance of the proposed method on the real collected point cloud in the KITTI dataset \cite{data_kitti}.
Figure \ref{fig:kitti} shows an example of the upsampling results.
Due to the hardware limitation of the LiDAR sensor, the collected point cloud is naturally sparse and non-uniformly distributed, making the upsampling more challenging.
Our method can generate dense point clouds with better distributions.

\subsection{Point Cloud Completion Task} \label{exp:3}
Then, we test the proposed upsampler on the point cloud completion task using the ShapeNet-PCN dataset.

\noindent
\textbf{ShapeNet-PCN:}
The ShapeNet-PCN dataset is introduced by \cite{pcn}, 
which is derived from ShapeNet \cite{chang2015shapenet}.
It contains pairs of partial and complete point clouds from 30,974 models of 8 categories in total: airplane, cabinet, car, chair, lamp, sofa, table, and watercraft. 
The complete point clouds are created by sampling 16,384 points uniformly from the original meshes, and the partial point clouds are generated by back-projecting 2.5D depth images into 3D. 
For each ground truth, 8 partial point clouds are generated from 8 randomly distributed viewpoints.
For fairness, we follow the same train/test splitting strategy in \cite{pcn,grnet,snowflakenet} with 29,774 training samples and 1,200 testing samples, and resample each incomplete cloud to 2,048 points.

\begin{figure}[bt]
	\centering
	\includegraphics[width=0.47\textwidth]{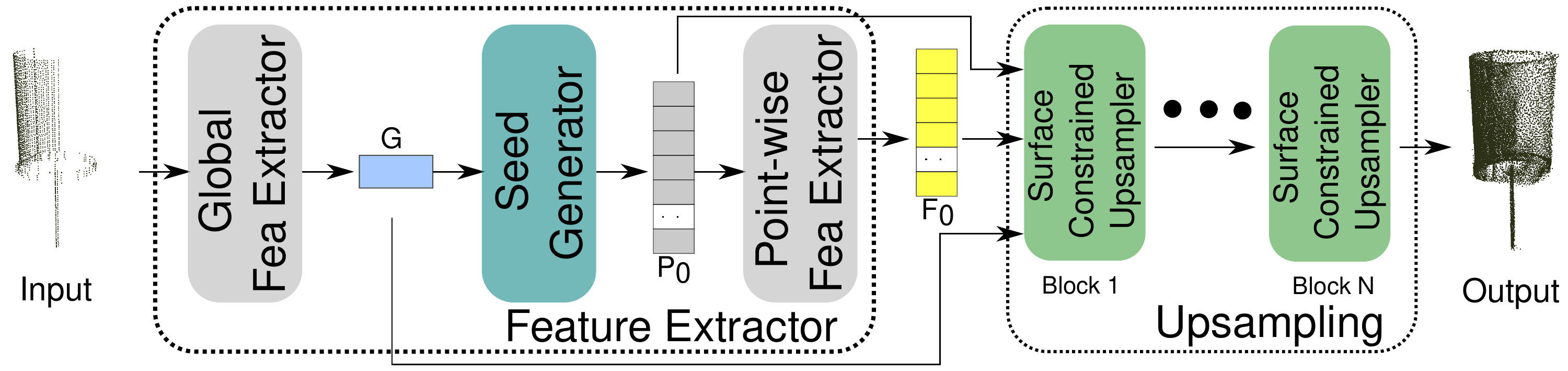}
	\caption{Network architecture for point cloud completion.}
	\label{fig:ShapeNet_network}
\end{figure}
\noindent
\textbf{Network Structure:}
As the point cloud completion task is more challenging,
where the input is a sparse and incomplete point cloud and the output is a dense and complete point cloud, we use a new network to generate complete and denser point clouds.
Figure \ref{fig:ShapeNet_network} shows the detailed network architecture for the point cloud completion task.
Because the input is incomplete, generating a coarse but complete point cloud is
crucial for subsequent upsampling steps.
Inspired by the previous SOTA algorithm SnowFlakeNet \cite{snowflakenet}, 
we adopted the seed generator used in their network to generate a sparse but complete point cloud,
then feed them into three consecutive surface constrained upsamplers with upscale ratios of 1, 4, and 8, respectively, to generate high resolution point clouds.

\noindent
\textbf{Evaluation Metrics:}
For a fair comparison with previous methods, we use two commonly used metrics:
L1 Chamfer Distance (L1-CD) and Earth Mover's Distance (EMD).
Similarly, the smaller the metric, the better the performance.

\noindent
\textbf{Training Detail:} We use 4 Tesla
V100 GPUs with a batch size of 32 and a total epoch number of 500.
Similar to SnowFlakeNet, we use Adam as the optimization function with warm-up settings, where it first takes 200 steps to warm up the learning rate from 0 to 0.0005,
and then the learning rate decays by a factor of 0.5 for every 50 epochs.

\begin{table}
	\centering
		\begin{tabular}{cccccc}
			\hline
			Methods & L1-CD ($\times 10^{-3}$) & EMD ($\times 10^{-3}$)\\
			\hline
            PCN  & 9.64 & 87.14\\
            GR-Net  & 8.83 & \bf{55.26}\\
            PMP  & 8.73 & 109.67 \\
            SnowFlake & \underline{7.19} & 69.13\\
            Ours & \bf{7.04} & \underline{66.57}\\
			\hline
		\end{tabular}
  \caption{Quantitative completion results compared to previous SOTA algorithms on the ShapeNet-PCN dataset.}
  \label{tab:shapenet}
\end{table}
\subsubsection{Experiment Results}
Table \ref{tab:shapenet} shows the quantitative completion results on the ShapeNet-PCN dataset.
We notice that our method still achieves the best performance in terms of L1-CD.
As we use the same backbone and upscale settings as SnowflakeNet, which is the previous SOTA algorithm, the improvement over SnowflakeNet can directly prove the effectiveness of our proposed upsampling blocks.
Compared to SnowflakeNet, we see that our network reduces the average L1-CD from $7.19 \times 10^{-3}$ to $7.04 \times 10^{-3}$ 
and the average EMD from $69.13 \times 10^{-3}$ to $66.57 \times 10^{-3}$.
Figure \ref{fig:shapenet} shows one completion result. Note that more completion results can be found in the Supplementary.
Still, we see that our method produces a much better shape quality and fewer outlier points.
\begin{figure}[bt]
	\centering
	\includegraphics[width=0.47\textwidth]{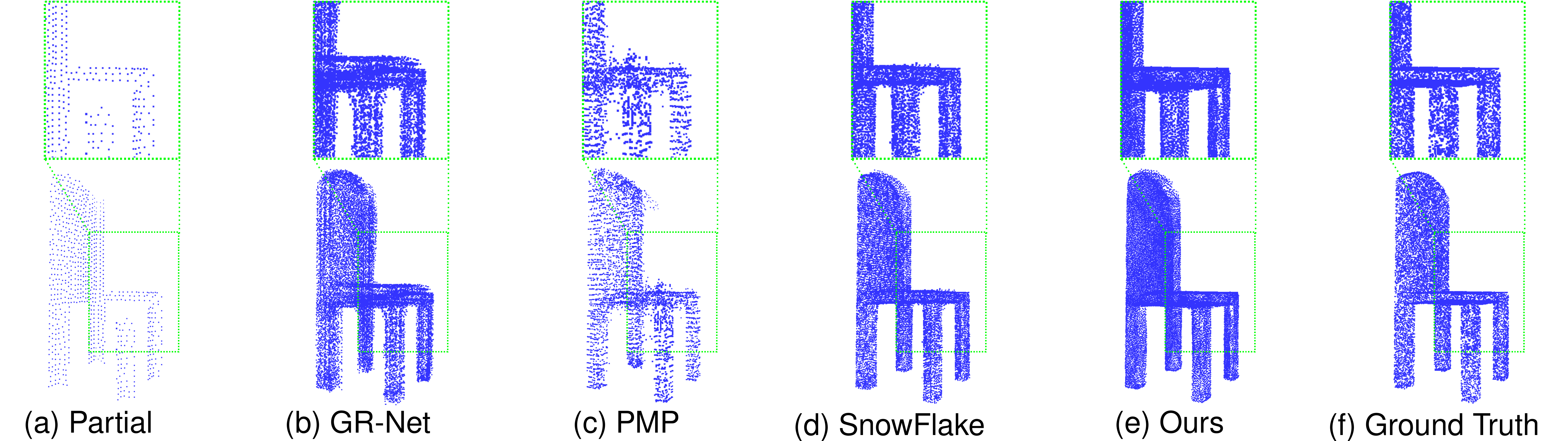}
	\caption{Visualization of completion results with different algorithms (GR-Net, PMP, SnowFlake-Net, and Ours).}
	\label{fig:shapenet}
\end{figure}

Note that our upsampler is designed based on the assumption that the input points are well-distributed. But for the completion task, this assumption is not met. Even under this challenging condition, the proposed upsampler still works and the generated points are still well-constrained to the underlying surface.

\section{Conclusions}
In this paper, we propose a novel parametric surface constrained upsampler for point clouds. 
By introducing explicit parametric surface functions into the network design, we can obtain better shape representation ability compared to MLPs with point-wise loss and generate point clouds with smoother shapes and fewer outliers.
In addition, the proposed upsampler can also be used in point completion tasks.
The experiment results on both point cloud upsampling and completion tasks prove the effectiveness of our method.

\section{Acknowledgements}
We sincerely thank the Senior Program Committee members and reviewers for their comments and contributions to the community. This work was supported, in part, by NEH PR-284350-22. The GPU used in this work was provided by the NSF MRI-2018966.

\bibliography{aaai23}

\begin{thebibliography}{48}
\providecommand{\natexlab}[1]{#1}

\bibitem[{Alexa et~al.(2003)Alexa, Behr, Cohen-Or, Fleishman, Levin, and
  Silva}]{traditional_1}
Alexa, M.; Behr, J.; Cohen-Or, D.; Fleishman, S.; Levin, D.; and Silva, C.
  2003.
\newblock Computing and rendering point set surfaces.
\newblock \emph{IEEE Transactions on Visualization and Computer Graphics},
  9(1).

\bibitem[{Bridson(2007)}]{poisson_disk}
Bridson, R. 2007.
\newblock {Fast Poisson Disk Sampling in Arbitrary Dimensions}.
\newblock In \emph{ACM SIGGRAPH 2007 Sketches}.

\bibitem[{Cai and Sur(2022)}]{DeepPCD}
Cai, P.; and Sur, S. 2022.
\newblock DeepPCD: Enabling AutoCompletion of Indoor Point Clouds with Deep
  Learning.
\newblock \emph{Proc. ACM Interact. Mob. Wearable Ubiquitous Technol.}, 6(2).

\bibitem[{Chang et~al.(2015)Chang, Funkhouser, Guibas, Hanrahan, Huang, Li,
  Savarese, Savva, Song, Su, Xiao, Yi, and Yu}]{chang2015shapenet}
Chang, A.~X.; Funkhouser, T.; Guibas, L.; Hanrahan, P.; Huang, Q.; Li, Z.;
  Savarese, S.; Savva, M.; Song, S.; Su, H.; Xiao, J.; Yi, L.; and Yu, F. 2015.
\newblock {ShapeNet: An Information-Rich 3D Model Repository}.
\newblock arXiv:1512.03012.

\bibitem[{do~Carmo(1976)}]{differential_geometry}
do~Carmo, M. 1976.
\newblock \emph{Differential geometry of curves and surfaces.}
\newblock Prentice Hall.
\newblock ISBN 978-0-13-212589-5.

\bibitem[{Engelmann et~al.(2017)Engelmann, Kontogianni, Hermans, and
  Leibe}]{Engelmann_2017_ICCV}
Engelmann, F.; Kontogianni, T.; Hermans, A.; and Leibe, B. 2017.
\newblock Exploring Spatial Context for 3D Semantic Segmentation of Point
  Clouds.
\newblock In \emph{IEEE International Conference on Computer Vision (ICCV)
  Workshops}.

\bibitem[{Fan, Su, and Guibas(2017)}]{FanSG17}
Fan, H.; Su, H.; and Guibas, L.~J. 2017.
\newblock A Point Set Generation Network for 3D Object Reconstruction from a
  Single Image.
\newblock In \emph{IEEE/CVF Conference on Computer Vision and Pattern
  Recognition (CVPR)}.

\bibitem[{Geiger et~al.(2013)Geiger, Lenz, Stiller, and Urtasun}]{data_kitti}
Geiger, A.; Lenz, P.; Stiller, C.; and Urtasun, R. 2013.
\newblock Vision meets Robotics: The KITTI Dataset.
\newblock \emph{International Journal of Robotics Research (IJRR)}, 32(11):
  1231--1237.

\bibitem[{Groueix et~al.(2018)Groueix, Fisher, Kim, Russell, and Aubry}]{atlas}
Groueix, T.; Fisher, M.; Kim, V.~G.; Russell, B.; and Aubry, M. 2018.
\newblock AtlasNet: A Papier-M\^ach\'e Approach to Learning 3D Surface
  Generation.
\newblock In \emph{IEEE/CVF Conference on Computer Vision and Pattern
  Recognition (CVPR)}.

\bibitem[{Huang et~al.(2013)Huang, Wu, Gong, Cohen-Or, Ascher, and Zhang}]{ear}
Huang, H.; Wu, S.; Gong, M.; Cohen-Or, D.; Ascher, U.; and Zhang, H.~R. 2013.
\newblock Edge-Aware Point Set Resampling.
\newblock \emph{ACM Transactions on Graphics}, 32(1).

\bibitem[{Huang et~al.(2020)Huang, Yu, Xu, Ni, and Le}]{PF-Net}
Huang, Z.; Yu, Y.; Xu, J.; Ni, F.; and Le, X. 2020.
\newblock PF-Net: Point Fractal Network for 3D Point Cloud Completion.
\newblock In \emph{IEEE/CVF Conference on Computer Vision and Pattern
  Recognition (CVPR)}.

\bibitem[{{Jolliffe, Ian}(2014)}]{PCA}
{Jolliffe, Ian}. 2014.
\newblock \emph{Principal Component Analysis}.
\newblock John Wiley \& Sons, Ltd.

\bibitem[{Landrieu and Simonovsky(2018)}]{Landrieu_2018_CVPR}
Landrieu, L.; and Simonovsky, M. 2018.
\newblock Large-Scale Point Cloud Semantic Segmentation With Superpoint Graphs.
\newblock In \emph{IEEE/CVF Conference on Computer Vision and Pattern
  Recognition (CVPR)}.

\bibitem[{Li et~al.(2019)Li, Li, Fu, Cohen-Or, and Heng}]{pugan}
Li, R.; Li, X.; Fu, C.-W.; Cohen-Or, D.; and Heng, P.-A. 2019.
\newblock PU-GAN: a Point Cloud Upsampling Adversarial Network.
\newblock In \emph{IEEE International Conference on Computer Vision (ICCV)}.

\bibitem[{Li et~al.(2020)Li, Li, Heng, and Fu}]{Li_2020_CVPR}
Li, R.; Li, X.; Heng, P.-A.; and Fu, C.-W. 2020.
\newblock PointAugment: An Auto-Augmentation Framework for Point Cloud
  Classification.
\newblock In \emph{IEEE/CVF Conference on Computer Vision and Pattern
  Recognition (CVPR)}.

\bibitem[{Li et~al.(2021)Li, Ma, Zhong, Liu, Chapman, Cao, and
  Li}]{pcdForAutoDrive2}
Li, Y.; Ma, L.; Zhong, Z.; Liu, F.; Chapman, M.~A.; Cao, D.; and Li, J. 2021.
\newblock Deep Learning for LiDAR Point Clouds in Autonomous Driving: A Review.
\newblock \emph{IEEE Transactions on Neural Networks and Learning Systems},
  32(8): 3412--3432.

\bibitem[{Lipman et~al.(2007)Lipman, Cohen-Or, Levin, and
  Tal-Ezer}]{traditional_2}
Lipman, Y.; Cohen-Or, D.; Levin, D.; and Tal-Ezer, H. 2007.
\newblock Parameterization-Free Projection for Geometry Reconstruction.
\newblock \emph{ACM Transactions on Graphics}, 26(3).

\bibitem[{Liu et~al.(2020)Liu, Sheng, Yang, Shao, and Hu}]{msn}
Liu, M.; Sheng, L.; Yang, S.; Shao, J.; and Hu, S.-M. 2020.
\newblock Morphing and sampling network for dense point cloud completion.
\newblock In \emph{AAAI conference on artificial intelligence}.

\bibitem[{Liu et~al.(2022)Liu, Liu, Liu, and Han}]{spu}
Liu, X.; Liu, X.; Liu, Y.-S.; and Han, Z. 2022.
\newblock SPU-Net: Self-Supervised Point Cloud Upsampling by Coarse-to-Fine
  Reconstruction with Self-Projection Optimization.
\newblock \emph{IEEE Transactions on Image Processing}, 31: 4213--4226.

\bibitem[{Long et~al.(2021)Long, Zhang, Li, Wang, Dong, and Yang}]{pc2pu}
Long, C.; Zhang, W.; Li, R.; Wang, H.; Dong, Z.; and Yang, B. 2021.
\newblock PC2-PU: Patch Correlation and Position Correction for Effective Point
  Cloud Upsampling.
\newblock arXiv:2109.09337.

\bibitem[{Luo and Hu(2020)}]{luo2020differentiable}
Luo, S.; and Hu, W. 2020.
\newblock Differentiable Manifold Reconstruction for Point Cloud Denoising.
\newblock In \emph{ACM International Conference on Multimedia}.

\bibitem[{Pan et~al.(2021)Pan, Chen, Cai, Zhang, Zhao, Yi, and Liu}]{vrpcn}
Pan, L.; Chen, X.; Cai, Z.; Zhang, J.; Zhao, H.; Yi, S.; and Liu, Z. 2021.
\newblock Variational Relational Point Completion Network.
\newblock arXiv:2104.10154.

\bibitem[{Preiner et~al.(2014)Preiner, Mattausch, Arikan, Pajarola, and
  Wimmer}]{traditional_5}
Preiner, R.; Mattausch, O.; Arikan, M.; Pajarola, R.; and Wimmer, M. 2014.
\newblock Continuous Projection for Fast L1 Reconstruction.
\newblock \emph{ACM Transactions on Graphics}, 33(4).

\bibitem[{Qi et~al.(2017{\natexlab{a}})Qi, Su, Mo, and Guibas}]{pointnet}
Qi, C.~R.; Su, H.; Mo, K.; and Guibas, L.~J. 2017{\natexlab{a}}.
\newblock Pointnet: Deep learning on point sets for 3d classification and
  segmentation.
\newblock In \emph{IEEE/CVF Conference on Computer Vision and Pattern
  Recognition (CVPR)}.

\bibitem[{Qi et~al.(2017{\natexlab{b}})Qi, Yi, Su, and
  Guibas}]{pointnetplusplus}
Qi, C.~R.; Yi, L.; Su, H.; and Guibas, L.~J. 2017{\natexlab{b}}.
\newblock Pointnet++: Deep hierarchical feature learning on point sets in a
  metric space.
\newblock In \emph{Advances in neural information processing systems}.

\bibitem[{Qian et~al.(2021)Qian, Abualshour, Li, Thabet, and Ghanem}]{PUGCN}
Qian, G.; Abualshour, A.; Li, G.; Thabet, A.; and Ghanem, B. 2021.
\newblock PU-GCN: Point Cloud Upsampling Using Graph Convolutional Networks.
\newblock In \emph{IEEE/CVF Conference on Computer Vision and Pattern
  Recognition (CVPR)}.

\bibitem[{Qian et~al.(2020)Qian, Hou, Kwong, and He}]{pugeo}
Qian, Y.; Hou, J.; Kwong, S.; and He, Y. 2020.
\newblock PUGeo-Net: A Geometry-Centric Network for 3D Point Cloud Upsampling.
\newblock In \emph{European Conference on Computer Vision (ECCV)}.

\bibitem[{Qiu, Anwar, and Barnes(2021)}]{PUTrans}
Qiu, S.; Anwar, S.; and Barnes, N. 2021.
\newblock PU-Transformer: Point Cloud Upsampling Transformer.
\newblock arXiv:2111.12242.

\bibitem[{Rusu et~al.(2008)Rusu, Marton, Blodow, Dolha, and
  Beetz}]{pcdForRobot}
Rusu, R.~B.; Marton, Z.~C.; Blodow, N.; Dolha, M.; and Beetz, M. 2008.
\newblock Towards 3D point cloud based object maps for household environments.
\newblock \emph{Robotics and Autonomous Systems}, 56(11): 927--941.

\bibitem[{Tchapmi et~al.(2019)Tchapmi, Kosaraju, Rezatofighi, Reid, and
  Savarese}]{top_net}
Tchapmi, L.~P.; Kosaraju, V.; Rezatofighi, H.; Reid, I.; and Savarese, S. 2019.
\newblock TopNet: Structural Point Cloud Decoder.
\newblock In \emph{IEEE/CVF Conference on Computer Vision and Pattern
  Recognition (CVPR)}.

\bibitem[{Wang, Ang, and Lee(2022)}]{CRN}
Wang, X.; Ang, M.~H.; and Lee, G. 2022.
\newblock Cascaded Refinement Network for Point Cloud Completion with
  Self-supervision.
\newblock \emph{IEEE Transactions on Pattern Analysis and Machine
  Intelligence}, 44(11): 8139--8150.

\bibitem[{Wang et~al.(2019)Wang, Wu, Huang, Cohen-Or, and
  Sorkine-Hornung}]{Wang2019PatchBasedP3}
Wang, Y.; Wu, S.; Huang, H.; Cohen-Or, D.; and Sorkine-Hornung, O. 2019.
\newblock Patch-Based Progressive 3D Point Set Upsampling.
\newblock In \emph{IEEE/CVF Conference on Computer Vision and Pattern
  Recognition (CVPR)}.

\bibitem[{Wen et~al.(2020)Wen, Li, Han, and Liu}]{pointSA}
Wen, X.; Li, T.; Han, Z.; and Liu, Y.-S. 2020.
\newblock Point Cloud Completion by Skip-Attention Network With Hierarchical
  Folding.
\newblock In \emph{IEEE/CVF Conference on Computer Vision and Pattern
  Recognition (CVPR)}.

\bibitem[{Wen et~al.(2021)Wen, Xiang, Han, Cao, Wan, Zheng, and Liu}]{pmp}
Wen, X.; Xiang, P.; Han, Z.; Cao, Y.-P.; Wan, P.; Zheng, W.; and Liu, Y.-S.
  2021.
\newblock PMP-Net: Point cloud completion by learning multi-step point moving
  paths.
\newblock In \emph{IEEE/CVF Conference on Computer Vision and Pattern
  Recognition (CVPR)}.

\bibitem[{Williams et~al.(2019)Williams, Schneider, Silva, Zorin, Bruna, and
  Panozzo}]{Deep_geometric_prior}
Williams, F.; Schneider, T.; Silva, C.; Zorin, D.; Bruna, J.; and Panozzo, D.
  2019.
\newblock Deep Geometric Prior for Surface Reconstruction.
\newblock In \emph{IEEE/CVF Conference on Computer Vision and Pattern
  Recognition (CVPR)}.

\bibitem[{Wu, Zhang, and Huang(2019)}]{wu2019point}
Wu, H.; Zhang, J.; and Huang, K. 2019.
\newblock Point Cloud Super Resolution with Adversarial Residual Graph
  Networks.
\newblock arXiv:1908.02111.

\bibitem[{Wu et~al.(2015)Wu, Huang, Gong, Zwicker, and
  Cohen-Or}]{traditional_4}
Wu, S.; Huang, H.; Gong, M.; Zwicker, M.; and Cohen-Or, D. 2015.
\newblock Deep Points Consolidation.
\newblock \emph{ACM Transactions on Graphics}, 34(6).

\bibitem[{Xiang et~al.(2021)Xiang, Wen, Liu, Cao, Wan, Zheng, and
  Han}]{snowflakenet}
Xiang, P.; Wen, X.; Liu, Y.-S.; Cao, Y.-P.; Wan, P.; Zheng, W.; and Han, Z.
  2021.
\newblock {SnowflakeNet}: Point Cloud Completion by Snowflake Point
  Deconvolution with Skip-Transformer.
\newblock In \emph{IEEE International Conference on Computer Vision (ICCV)}.

\bibitem[{Xie et~al.(2020)Xie, Yao, Zhou, Mao, Zhang, and Sun}]{grnet}
Xie, H.; Yao, H.; Zhou, S.; Mao, J.; Zhang, S.; and Sun, W. 2020.
\newblock GRNet: Gridding Residual Network for Dense Point Cloud Completion.
\newblock In \emph{European Conference on Computer Vision (ECCV)}.

\bibitem[{Yang et~al.(2018)Yang, Feng, Shen, and Tian}]{foldingnet}
Yang, Y.; Feng, C.; Shen, Y.; and Tian, D. 2018.
\newblock Foldingnet: Point cloud auto-encoder via deep grid deformation.
\newblock In \emph{IEEE/CVF Conference on Computer Vision and Pattern
  Recognition (CVPR)}.

\bibitem[{Ye et~al.(2022)Ye, Chen, Han, Wan, and Liao}]{MetaPUAA}
Ye, S.; Chen, D.; Han, S.; Wan, Z.; and Liao, J. 2022.
\newblock Meta-PU: An Arbitrary-Scale Upsampling Network for Point Cloud.
\newblock \emph{IEEE Transactions on Visualization and Computer Graphics},
  28(9): 3206--3218.

\bibitem[{Yu et~al.(2018)Yu, Li, Fu, Cohen-Or, and Heng}]{punet}
Yu, L.; Li, X.; Fu, C.-W.; Cohen-Or, D.; and Heng, P.-A. 2018.
\newblock PU-Net: Point Cloud Upsampling Network.
\newblock In \emph{IEEE/CVF Conference on Computer Vision and Pattern
  Recognition (CVPR)}.

\bibitem[{Yuan et~al.(2018)Yuan, Khot, Held, Mertz, and Hebert}]{pcn}
Yuan, W.; Khot, T.; Held, D.; Mertz, C.; and Hebert, M. 2018.
\newblock PCN: Point Completion Network.
\newblock In \emph{International Conference on 3D Vision (3DV)}.

\bibitem[{Zeng et~al.(2018)Zeng, Hu, Liu, Ye, Han, Li, and
  Sun}]{pcdForAutoDrive}
Zeng, Y.; Hu, Y.; Liu, S.; Ye, J.; Han, Y.; Li, X.; and Sun, N. 2018.
\newblock {RT3D: Real-Time 3-D Vehicle Detection in LiDAR Point Cloud for
  Autonomous Driving}.
\newblock \emph{IEEE Robotics and Automation Letters}, 3(4): 3434--3440.

\bibitem[{Zhang et~al.(2020)Zhang, Fang, Wah, and Torr}]{pcd_seg}
Zhang, F.; Fang, J.; Wah, B.; and Torr, P. 2020.
\newblock Deep FusionNet for Point Cloud Semantic Segmentation.
\newblock In \emph{European Conference on Computer Vision (ECCV)}.

\bibitem[{Zhang, Yan, and Xiao(2020)}]{SFA}
Zhang, W.; Yan, Q.; and Xiao, C. 2020.
\newblock Detail Preserved Point Cloud Completion via Separated Feature
  Aggregation.
\newblock In \emph{European Conference on Computer Vision (ECCV)}.

\bibitem[{Zhao et~al.(2021)Zhao, Jiang, Jia, Torr, and
  Koltun}]{point_transformer}
Zhao, H.; Jiang, L.; Jia, J.; Torr, P.~H.; and Koltun, V. 2021.
\newblock Point transformer.
\newblock In \emph{IEEE/CVF Conference on Computer Vision and Pattern
  Recognition (CVPR)}.

\bibitem[{Zhou et~al.(2019)Zhou, Barnes, Jingwan, Jimei, and
  Hao}]{Rotation_Representations}
Zhou, Y.; Barnes, C.; Jingwan, L.; Jimei, Y.; and Hao, L. 2019.
\newblock On the Continuity of Rotation Representations in Neural Networks.
\newblock In \emph{IEEE/CVF Conference on Computer Vision and Pattern
  Recognition (CVPR)}.

\end{thebibliography}

\twocolumn[{%
	\renewcommand\twocolumn[1][]{#1}%
	\maketitle
	\begin{center}
		\centering
		\captionsetup{type=figure}
		\includegraphics[width=1\textwidth]{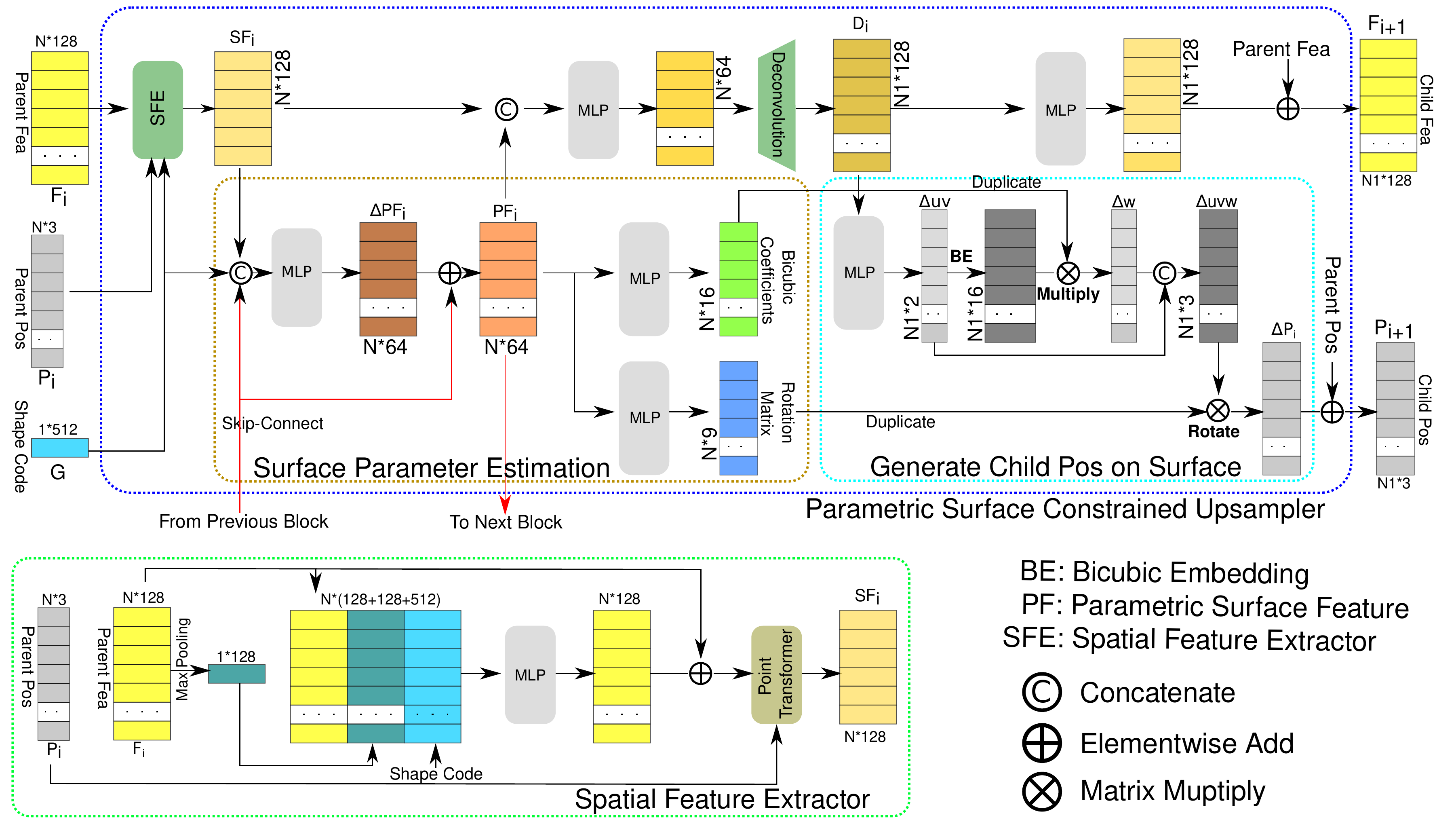}
		\captionof{figure}{The detailed network architecture for parametric surface constrained upsampler, with three major components: Spatial Feature Extractor (SFE), Surface Parameter Estimation, and Generate Child Points on Surface.}
		\label{fig:full_network}
	\end{center}%
}]
\section{Detailed Network Design and Hyperparameter}
\begin{figure*}[hbt]
	\vspace{0mm}
	\centering
	\includegraphics[width=1\textwidth]{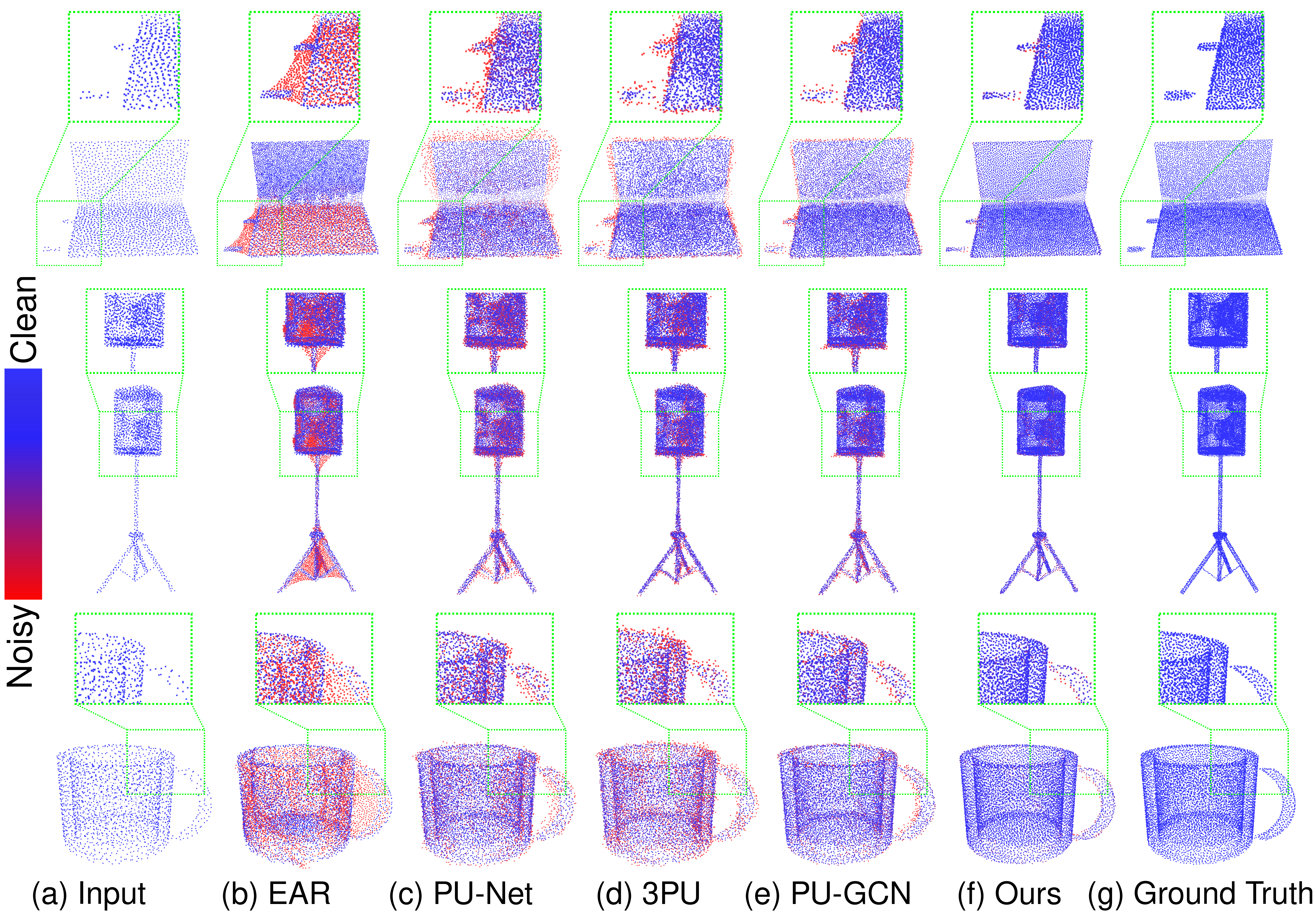}
	\caption{Visualization of upsampling results on the PU1K dataset with different algorithms (EAR (\citeauthor{ear}), PU-Net (\citeauthor{punet}), 3PU (\citeauthor{Wang2019PatchBasedP3}), PUGCN (\citeauthor{PUGCN}), and Ours). Please enlarge the PDF for more details.}
	\label{fig:upsampling}
	\vspace{-1mm}
\end{figure*}
\noindent
Figure \ref{fig:full_network} shows the detailed network architecture of the proposed upsampler with specific dimensions for each intermediate feature map,
where the dimension of the feature maps $C_1$ and $C_2$ are set to 128 and 512 arbitrarily and $N1=m*N$.
Note that other settings are also feasible.
In the spatial feature extractor, we set the number of nearest neighbors $K=16$ for the point transformers (\citeauthor{point_transformer}) to aggregate the context of neighbor points.
In the surface parameter estimation module,
we incorporate the global shape code G to smooth the spatial feature
$SF_i$ via an MLP and generate the smoothed Parametric surface Feature $PF_{i} \in R^{N \times 64}$, which will be used to predict the bicubic coefficients and rotation matrix via MLPs.
In addition, since the local surface around each child point can be viewed as a refined subsurface of its parent surface, we can also add a skip connection between two consecutive upsamplers to facilitate training.

In training, we set the hyperparameter $\lambda$ in the loss function to $1$ to balance the Chamfer Distance Loss and the proposed displacement loss.

\section{More Experiment Results}

\subsection{Point Cloud Upsampling}
We first show more visual results on the point-cloud upsampling task.
Figure \ref{fig:upsampling} illustrates additional upsampling results on the PU1K dataset with a variety of complex objects.
We see that the traditional algorithm EAR (\citeauthor{ear}) fails to preserve the geometric structure of the input point clouds even when the input shape is simple and outperformed by deep learning-based algorithms. 
On the other hand, previous deep learning algorithms fail to constrain the upsampled points and 
tend to generate noisy and outlier points. 
By using the proposed parametric surface constrained upsampler, we can generate much better point clouds with accurate shapes.
For example, in the case of the cup, the upsampled points near the handle are much cleaner than others, with fewer off-the-surface points. 
Similar results can be found in other objects. These visual results further validate the effectiveness of the proposed upsampler in the point cloud upsampling task, 
and the generated child points preserve better underlying surface properties and shape.
\begin{table*}[hbt]
	\vspace{-0mm}
	\begin{center}
		\begin{tabular}{c|c|cccccccc}
			\hline\noalign{\smallskip}
			Methods & Average & Plane & Cabinet & Car & Chair & Lamp & Couch & Table & Watercraft \\
			\noalign{\smallskip}
			\hline
			\noalign{\smallskip}
			FoldingNet (\citeauthor{foldingnet}) & 14.31  & 9.49 & 15.80 & 12.61 & 15.55 & 16.41 & 15.97 & 13.65 & 14.99\\
			TopNet (\citeauthor{top_net}) & 12.15 & 7.61 & 13.31 & 10.90 & 13.82 & 14.44 & 14.78 & 11.22 & 11.12 \\
			AtlasNet (\citeauthor{atlas}) & 10.85 & 6.37 & 11.94 & 10.10 & 12.06 &  12.37 & 12.99 & 10.33 & 10.61\\
			PCN (\citeauthor{pcn})  & 9.64 &5.50  & 22.70 &10.63 & 8.70 &11.0 & 11.34&11.68&8.59 \\
			GR-Net (\citeauthor{grnet}) & 8.83 & 6.45  & 10.37 & 9.45 & 9.41 &7.96 &10.51 & 8.44& 8.04 \\
			PMP (\citeauthor{pmp}) & 8.73 & 5.65&11.24&9.64&9.51&6.95&10.83&8.72&7.25\\
			CDN (\citeauthor{CRN}) & 8 .51 & 4.79 &9.97 & 8.31 &9.49 &8.94 &10.69 &7.81 &8.05 \\
			NSFA (\citeauthor{SFA}) & 8.06 & 4.76 & 10.18& 8.63 & 8.53 & 7.03 &10.53 &7.35 &7.48 \\
			SnowFlake (\citeauthor{snowflakenet}) & 7.19&4.24&9.27&8.20&7.75&\textbf{5.96}&9.25&6.45&6.37 \\
			Ours &\textbf{7.04} & \textbf{4.10}&\textbf{9.08}&\textbf{7.94}&\textbf{7.64}&6.07&\textbf{8.96}&\textbf{6.25}&\textbf{6.27}\\
			\hline
		\end{tabular}
		\caption{Point cloud completion results compared to previous algorithms (L1-CD $\times 10^{-3}$).}
		\label{tab:shapenet}
	\end{center}
	\vspace{0mm}
\end{table*}

\begin{figure*}[hbt]
	\vspace{-2mm}
	\centering
	\includegraphics[width=1\textwidth]{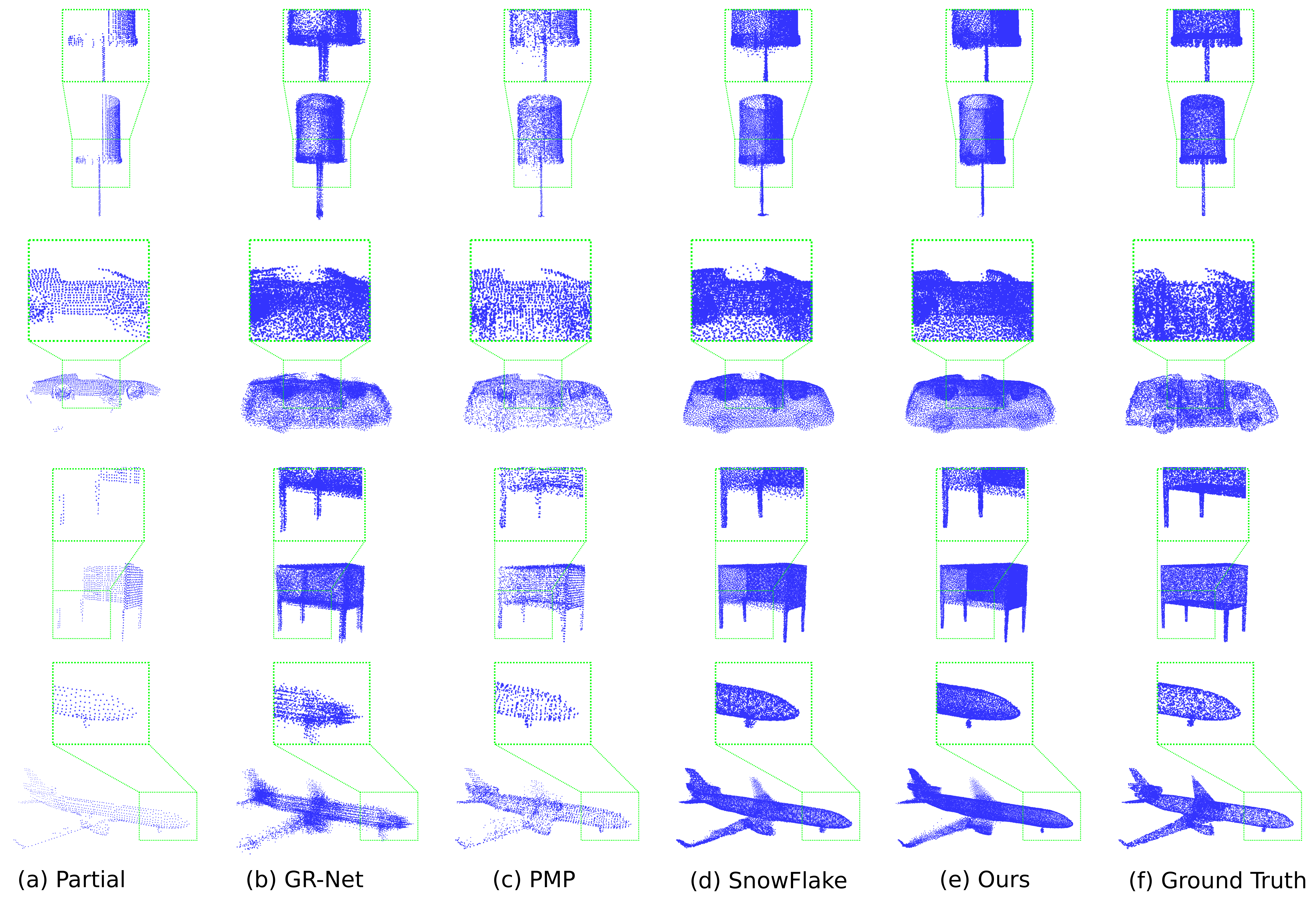}
	\caption{Visualization of completion results on ShapeNet-PCN dataset with different algorithms. We see that our method can generate points that are closer to the underlying surfaces, preserving better shapes.
		Please enlarge the PDF for more details.}
	\label{fig:completion}
	\vspace{0mm}
\end{figure*}

\subsection{Point Cloud Completion}
We then show more detailed results on the point cloud completion task.
Table \ref{tab:shapenet} shows the quantitative results of the ShapeNet-PCN dataset with detailed performance for each category compared to other methods.
Note that more methods are included and their results are directly cited from SnowFlakeNet (\citeauthor{snowflakenet}).
Compared to the previous SOTA methods,
we see that the proposed method achieves the best performance in 7 categories, showing good generalization ability.
What's more, as we use the same backbone and upscale settings in SnowflakeNet (\citeauthor{snowflakenet}), the improvement over it can directly prove the effectiveness of the proposed upsampler.
Figure \ref{fig:completion} shows more point cloud completion results in the ShapeNet-PCN dataset.
We see that the proposed upsampler can generate much better point clouds with accurate shapes.
Taking the desk as an example, the point distribution near the surface is smoother, with fewer off-the-surface points compared to others. The same observation can also be found in the car and aircraft, where the left door of the car and the front wheel of the aircraft are cleaner and much more similar to the ground truth. 
These evaluation results prove that, even for the challenging completion task, the generated points still preserve better underlying surface properties and shape by using the proposed surface constrained upsampler.

\end{document}